\documentclass[sigconf]{acmart}

\AtBeginDocument{%
  \providecommand\BibTeX{{%
    \normalfont B\kern-0.5em{\scshape i\kern-0.25em b}\kern-0.8em\TeX}}}

\copyrightyear{2021}
\acmYear{2021}
\setcopyright{acmcopyright}\acmConference[CIKM '21]{Proceedings of the 30th ACM
International Conference on Information and Knowledge Management}{November
1--5, 2021}{Virtual Event, QLD, Australia}
\acmBooktitle{Proceedings of the 30th ACM International Conference on Information
and Knowledge Management (CIKM '21), November 1--5, 2021, Virtual Event, QLD,
Australia}
\acmPrice{15.00}
\acmDOI{10.1145/3459637.3482082}
\acmISBN{978-1-4503-8446-9/21/11}

\acmSubmissionID{rgsp2629}

\usepackage{multirow}
\usepackage{array}
\usepackage{fancyhdr}
\begin{document}
\fancyhead{}
\title{Density-Based Dynamic Curriculum Learning \\for Intent Detection}

\author{
  Yantao Gong$^{1,2,3}$, 
  Cao Liu$^3$, 
  Jiazhen Yuan$^1$, 
  Fan Yang$^3$, \\
  Xunliang Cai$^3$, 
  Guanglu Wan$^3$, 
  Jiansong Chen$^3$, 
  Ruiyao Niu$^3$, 
  Houfeng Wang$^2$ \\
  $^1$ School of Software and Microelectronics, Peking University, 
  $^2$ MOE Key Lab of Computational Linguistics, Peking University,
  $^3$ Meituan,
  \{gongyt, yuanjiazhen, wanghf\}@pku.edu.cn
  \{liucao, yangfan79, caixunliang, wanguanglu, chenjiansong, niuruiyao\}@meituan.com
}

\begin{abstract}
  Pre-trained language models have achieved noticeable performance on the intent detection task. 
  However, due to assigning an identical weight to each sample, they suffer from the overfitting of simple samples and the failure to learn complex samples well. 
  To handle this problem, we propose a density-based dynamic curriculum learning model. 
  Our model defines the sample's difficulty level according to their eigenvectors' density. In this way, we exploit the overall distribution of all samples' eigenvectors simultaneously. 
  Then we apply a dynamic curriculum learning strategy, which pays distinct attention to samples of various difficulty levels and alters the proportion of samples during the training process. 
  Through the above operation, simple samples are well-trained, and complex samples are enhanced. 
  Experiments on three open datasets verify that the proposed density-based algorithm can distinguish simple and complex samples significantly. 
  Besides, our model obtains obvious improvement over the strong baselines. 
\end{abstract}





\maketitle

{
\medskip\small\noindent{\bfseries ACM Reference Format:}\par\nobreak
\noindent\bgroup\def\\{\unskip{}, \ignorespaces}{Yantao Gong, Cao Liu, Jiazhen Yuan, Fan Yang, Xunliang Cai, Guanglu Wan, Jiansong Chen, Ruiyao Niu, Houfeng Wang}\egroup. 2021. Density-Based Dynamic Curriculum Learningfor Intent Detection. In \textit{Proceedings of the 30th ACM Int'l Conf. on Information and Knowledge Management (CIKM '21), November 1--5, 2021, Virtual Event, QLD, Australia}\textit{.} ACM, New York, NY, USA, \ref{TotPages}~pages. https://doi.org/10.1145/3459637.3482082
}

\section{Introduction}
Intent detection is a crucial portion in understanding user queries, 
which usually predicts intent tags by semantic classification. 
It is widely used in search, task-based dialogue, and other fields \cite{brenes2009survey}. 

\begin{figure}[]
  \centering
    \includegraphics[width=\linewidth]{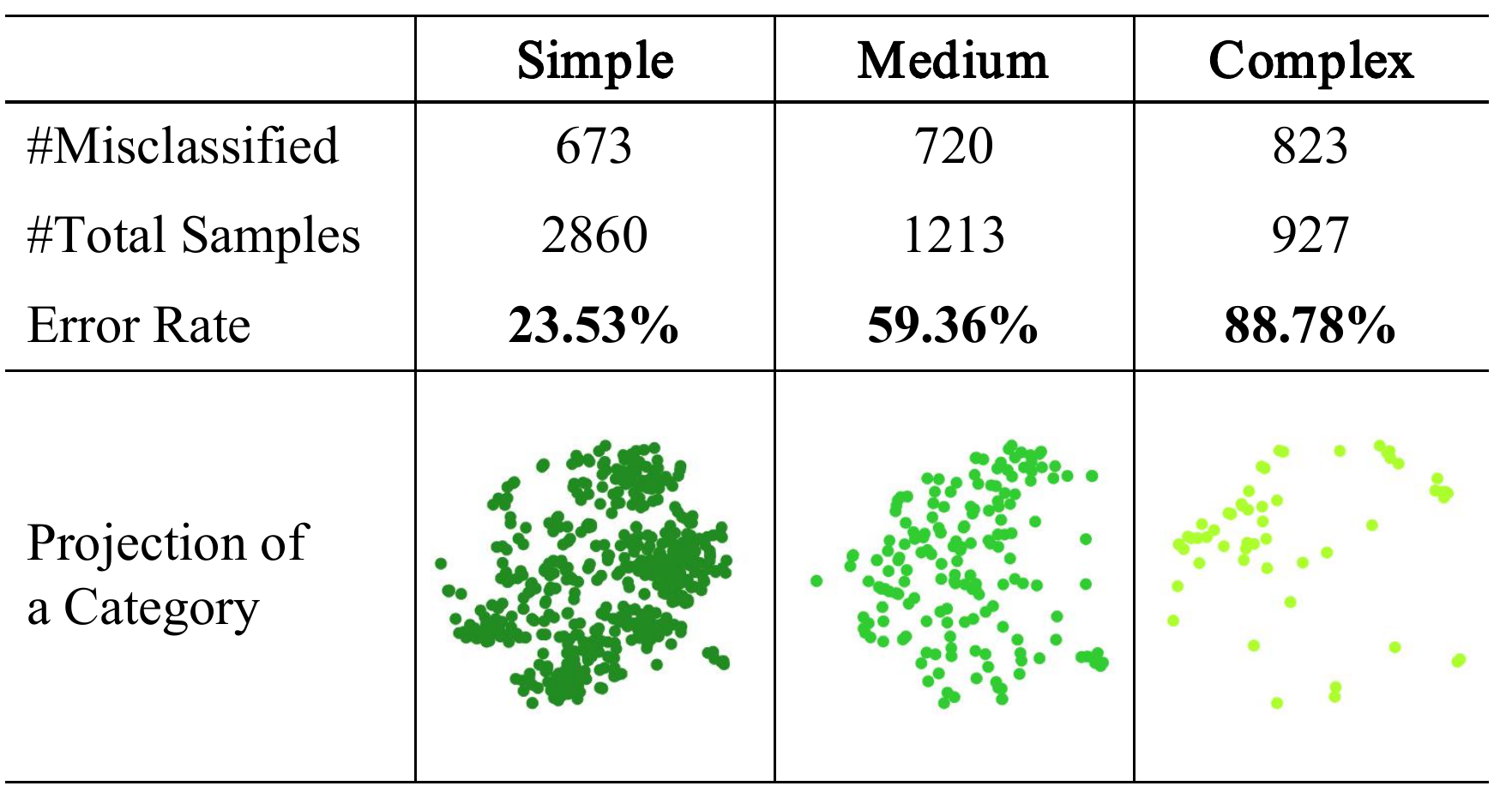}
    \caption{
      Eigenvector distribution of samples with various difficulty levels from 
      TNEWS dataset, using t-SNE algorithm for visualization. 
    Simple samples have a much lower error rate compared to complex samples. Besides, the eigenvectors of simple samples are in a higher density,   
    while the eigenvectors of complex samples are in a lower density. 
    }
    \Description{Enjoying the baseball game from the third-base
    seats. Ichiro Suzuki preparing to bat.}
    \label{fig:density}
\end{figure}

Recently, with the development of pre-trained language models, BERT-based models have achieved remarkable enhancement on 
multiple intent detection corpora. Nevertheless, there exist 
some major drawbacks yet. For instance, all samples are treated equally, leading 
to the overfitting of simple samples and barely satisfactory process on complex samples. 
As demonstrated in Figure \ref{fig:density}, we use BERT as the classifier on a real dataset. 
Complex samples have a much higher error rate than simple samples. 

One of the effective solutions to address the aforementioned problems is curriculum learning \cite{bengio2009curriculum}, as imitated 
from the bionic method of human learning. 
Curriculum learning can better guide the model to make the utmost of samples with various difficulty levels, by the paradigm that learns from simple to complex. 
Hence, the model is capable of achieving a more preferable outcome than conventional approaches.  

In fact, there already have several works on how to introduce curriculum 
learning into NLP tasks \cite{liu2020norm, liu2018curriculum}, which achieve satisfying performance. 
The key of curriculum learning is to define the sample's difficulty level. 
For instance, \cite{xu2020curriculum} 
uses multiple pre-trained sub-models to evaluate the difficulty of samples. However, these curriculum 
learning models suffer from the following issues: 
(1) They focus on the independent information of a single sample to define samples' difficulty, such as classification result of correct or wrong, 
without considering the overall distribution of all samples.  
(2) The capability of sub-models to define simple or complex samples is fixed from the initial 
phase and cannot be dynamically updated. 
However, 
the model's capability is actually different during various training stages.

Fortunately, we find that the eigenvectors extracted from samples are of different density distribution. 
Moreover, there is a corresponding relation between the complexity and eigenvectors' density. 
Consequently, the overall distribution of all samples could be exploited rather than just a single sample's information like \cite{xu2020curriculum}. 
We separate samples from TNEWS dataset into three difficulty levels,  
project their eigenvectors into two-dimensional space. 
As shown in Figure \ref{fig:density}, simple samples have a higher density of 
eigenvectors, while more complex samples have a lower density. 

Based on the above observations, we propose a density-based dynamic curriculum 
learning model, which contains two parts as below. 
(1) We use the distribution density of sample's eigenvector to 
define the difficulty level. In this way, samples' difficulty can be considered from the interaction of all samples in the entire spatial distribution, 
rather than the difficulty of each sample alone, like \cite{xu2020curriculum}. 
(2) After defining the difficulty of samples, the dynamic curriculum learning 
strategy changes the proportion of samples with various difficulty levels. New eigenvectors are 
extracted according to the model's current capability, which is updated in the training process other than remains constant like \cite{liu2020norm}. 


To verify the effectiveness of our model, we conduct abundant experiments on three open datasets. 
Empirical results show that exploiting the eigenvector distribution density can distinguish the simple and complex samples significantly. 
Combined with a dynamic curriculum learning strategy, the proposed model obtains evident improvement compared with the strong baselines. 


\section{Related Work}
\textbf{Intent detection. }
Intent is the purpose of an utterance such as a query generated by users. 
Actually, the essence of intent detection is text classification. 
Given a labeled training set, the model tries to predict the intent of a query in the existing intent sets. 
Massive researches have been done 
during the past few decades \cite{brenes2009survey, Wang2020MaskedfieldPF} for intent detection. 

\noindent \textbf{Curriculum learning. } CL has acquired striking results in lots of fields \cite{Ferro2018ContinuationMA, Penha2020CurriculumLS, guo2018curriculumnet}.  
Defining the samples' difficulty is the essential step in curriculum learning. 
In the machine translation task of NLP, 
\cite{liu2020norm} exploits the norm of words' vectors to define the samples' difficulty, where frequent words and context-insensitive words have smaller norms. 
Concerning QA generation, \cite{liu2018curriculum} proposed a model to make full use of the complexity and quality of QA pairs to generate a natural answer. 

\section{Method}


\subsection{Model Overview}

\begin{figure}[h]
  \centering
    \includegraphics[width=\linewidth]{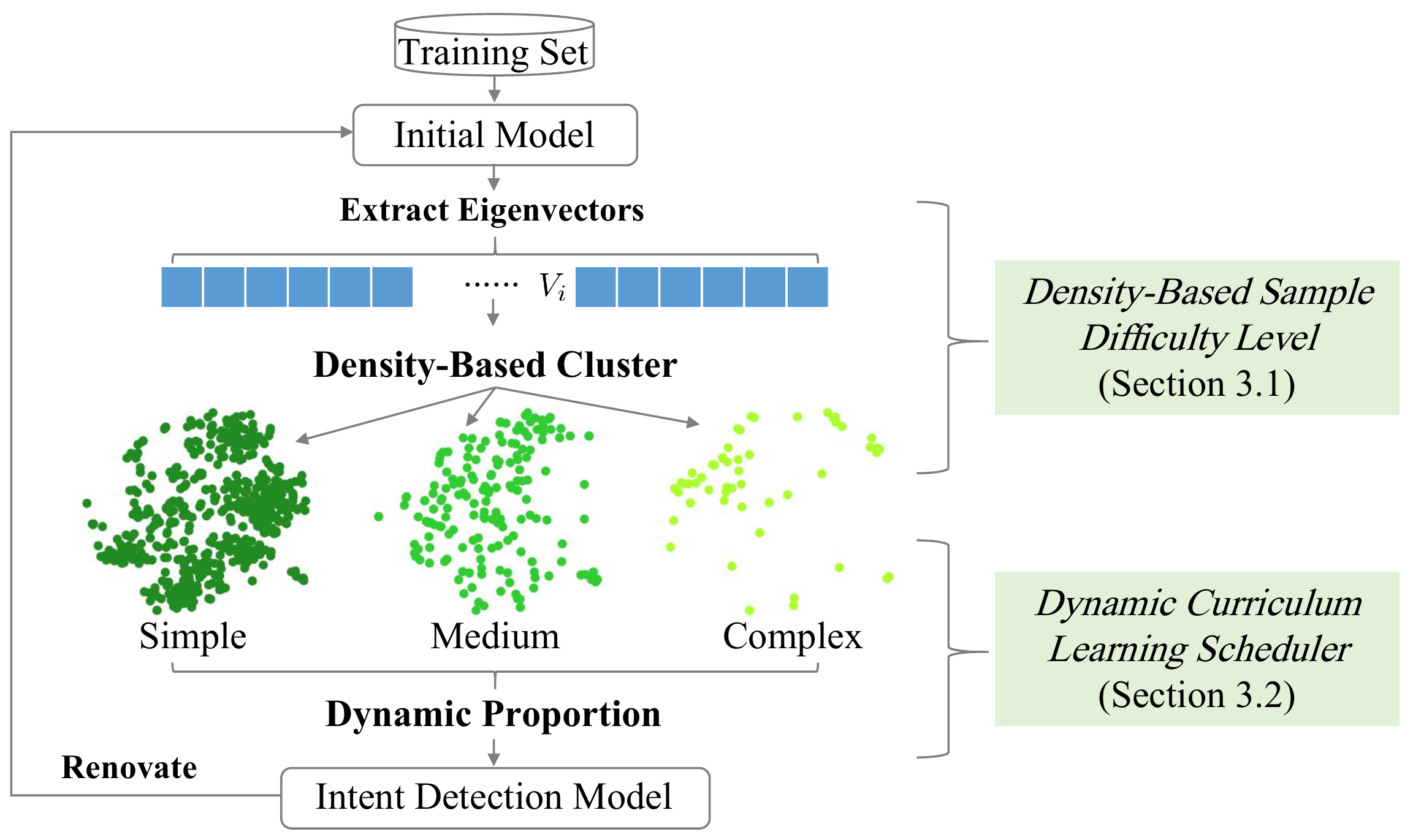}
    \caption{The illustration of \textit{density-based dynamic curriculum learning} framework. 
    We employ the initial model to obtain sample's eigenvector,  
    and exploit the density-based cluster algorithm to define the difficulty level of all samples simultaneously. 
    Then, we use dynamic curriculum learning to train the intent detection model, which replaces the initial model and renovates the capability for definition. 
    }
    \Description{Enjoying.}
    \label{fig:model}
\end{figure}

Exhibited in Figure \ref{fig:model},  
we use the initial model such as original BERT to extract the encoded vector from the training set. 
Through the above preprocess, we acquire the eigenvector of each sample. 
Next, the distribution density of eigenvectors is leveraged to cluster and define the difficulty level of sample. 
After that, we employ a dynamic curriculum 
learning scheduler. It changes the proportion of simple and complex samples, concurrently, pays distinct attention to various samples. 
Then, we get the intent detection model, which replaces the initial model and renovates the capability of defining difficulty level at various training stages. 
Last, repeating the operation until the end of training. 

\subsection{Density-Based Sample Difficulty Level}
\label{sec:density} 
In this submodule, we define the sample's difficulty level by the density of eigenvectors. 
Suppose a query $Q_i$ of sample $S_i$. The corresponding category is $C_t$. We exploit the initial model like BERT to extract the encoded vector $V_i$ of query $Q_i$, such as [CLS], which is defined as the eigenvector of sample. 
Take category $C_t$ as an example, 
we calculate Manhattan distance $d_{ij}$ between samples within 
the same category $C_t$ by the following: 
\begin{equation}
  d_{ij}= {\left | V_{i}-V_{j} \right |}
\end{equation}
where $V_i$ is the eigenvector of sample $S_i$, 
$V_j$ is the eigenvector of another sample $S_j$ in the identical category $C_t$. 

\begin{table*}
  \caption{Comparison on three open datasets, with or without curriculum learning.}
  \label{tab:main}
  \begin{tabular}{lcccccccccccc}
  \toprule[0.9pt]
  \multicolumn{1}{c}{\multirow{2}{*}{\textbf{Model}}} & \multicolumn{4}{c}{\textbf{TNEWS}}                               & \multicolumn{4}{c}{\textbf{BANKING77}}                               & \multicolumn{4}{c}{\textbf{CLINC150}}                                \\ \cline{2-13}
  \multicolumn{1}{c}{}                       & \textbf{P}     & \textbf{R}     & \textbf{F1}    & \textbf{Acc} & \textbf{P}     & \textbf{R}     & \textbf{F1}    & \textbf{Acc} & \textbf{P}     & \textbf{R}     & \textbf{F1} & \textbf{Acc} \\ \midrule[0.9pt]
  BERT                                  & 55.85          & 55.62          & 55.58       & 55.63             & 93.60          & 93.31          & 93.30          & 93.30             & 97.08          & 96.83          & 96.82          & 96.80             \\
  BERT+CL (\textbf{Ours})                              & \textbf{56.84} & \textbf{56.55} & \textbf{56.50}       & \textbf{56.55}    & \textbf{93.95} & \textbf{93.70} & \textbf{93.71} & \textbf{93.70}    & \textbf{97.26} & \textbf{97.07} & \textbf{97.04} & \textbf{97.07}    \\ \hline
  RoBERTa                               & 56.85          & 56.68          & 56.62       & 56.68             & 93.96          & 93.70          & 93.69          & 94.10             & 97.20          & 97.00          & 96.99          & 97.00             \\
  RoBERTa+CL (\textbf{Ours})                           & \textbf{57.70} & \textbf{57.23} & \textbf{57.30}       & \textbf{57.23}    & \textbf{94.38} & \textbf{94.12} & \textbf{94.13} & \textbf{94.12}    & \textbf{97.44} & \textbf{97.23} & \textbf{97.22} & \textbf{97.23}    \\ \bottomrule[0.9pt]
  \end{tabular}
  \end{table*}
  
After that, we obtain an array of distance \{$d_{12}, ..., d_{ij}, ..., d_{n(n-1)}$\} between eigenvectors within category $C_t$. 
$n$ is the total number of samples in category $C_t$. 
After sorting the above array in ascending order, we select the distance value ranked at $\theta \%$ 
to be the demarcation flag $d_{flag}$. Threshold $\theta$ is a hyperparameter set as 60. For each sample 
in category $C_t$, if the distance $d_{ij}$ between it and another sample in the same category $C_t$ 
is less than the demarcation flag $d_{flag}$, the density value is accumulated by one. 
The samples in this category eventually get their own accumulated density value through the equation as below: 

\begin{equation}
\label{equation:equation_density}
D_i={\textstyle \sum_{j=1}^{n}} Z_{ij}, \ \ \   ~~~~~~~~Z_{ij} = \begin{cases}
  1,  & \text{ if } d_{ij} < d_{flag} \\
  0,  & \text{ otherwise. }
  \end{cases}
\end{equation}
where $Z_{ij}$ is the accumulated value between sample $S_i$ and $S_j$, $j\ne i$. $D_i$ is the density value of sample $S_i$. 
Based on the density values of the samples in the category $C_t$, 
we exploit Kmeans to divide the samples into $K$ difficulty levels \{1, 2, ..., K\}. $K$ is a hyperparameter. Samples of each level are paid different attention in dynamic curriculum learning scheduler. 
Samples that have close density values are more likely to be clustered together. Therefore, 
samples with higher density values are clustered and defined as simple samples.  
In the meantime, complex samples are those with lower density.

\subsection{Dynamic Curriculum Learning Scheduler}
\label{sec:dynamic}

After acquiring the difficulty level of each sample, we design a dynamic curriculum learning strategy. To better utilize both simple and complex samples as well as pay different attention to them, we change the proportion and 
calculate the attention weight $\omega _k$ during each epoch as: 
\begin{equation}
  \omega _k = f_k(\lambda^{(-epoch_l)}),   \ \ \    k\in\{1, 2, ..., K\}  
\end{equation}
where $epoch_l$ is the number of current epoch, $f_k$ is the scheduler function, $k$ is the exact difficulty level of sample, $k=K$ represents the difficulty level of most complex samples. 
We exploit ${\lambda^{(-epoch_l)}}$, which makes a narrower 
fluctuation margin of the sample number in the later stage of training. 
$\lambda$ is a hyperparameter. 
Then the new number (${Num}'_k$) of samples in difficulty level $k$ for training is calculated by: 
\begin{equation}
  {Num}'_k = \omega _k * Num_k   
\end{equation}
where $Num_k$ is the original number of samples. 
As training progressing, simple samples are gradually reduced, 
while complex samples are selected. Moreover, suppose the number of complex samples is not reduced compared to the previous round. 
In this case, the number of different samples remains the same as the previous round, which encourages the model to find a better partition of training set containing fewer complex samples. 
So that models' capability of defining difficulty level is updated at each round. 


\section{Experiments}

\subsection{Experimental Setup}

\textbf{Datasets. }We mainly experiment on three open datasets, TNEWS, BANKING77, and CLINC150. 
TNEWS, proposed by \cite{xu2020clue}, has identical essence with intent detection. It includes 53360 samples in 15 categories. The provided test set are without gold labels. So we regard the validation set as the test set and randomly 
divide 5000 samples from the train set for validation. 
BANKING77, proposed by \cite{casanueva2020efficient}, has 13083 samples. It is the inquire dataset of online banking, consists of 77 intents in a single domain. 
CLINC150 proposed by \cite{larson2019evaluation}, contains 150 intents of 10 domains and 22500 total samples. 

\textbf{Implementation Details. }For baseline methods, we use BERT-based models as strong competitors. 
Then, we reproduce the baselines, and the results are almost equal to the published metrics. 
We set the difficulty level $K=3$ unless otherwise specified. 
Besides, we employ Adam as the optimizer and search learning rate in \{2e-5, 3e-5\}, with epochs in \{15, 19\}. To make full use of GPU memory, we set batch size equal to 256. 
The type of GPU is Tesla V100. 

\subsection{Overall Comparison}

\textbf{Comparison Settings.} In order to verify the effectiveness of our model, extensive 
experiments have been conducted on the above three datasets. 
We choose BERT and RoBERTa as the baselines for analysis, and regard the accuracy as the main evaluation metrics, Precision (P), Recall (R), and F1 for comparison as well. 
  
\textbf{Comparison Results.} As illustrated in Table \ref{tab:main}, we clearly see the following observations: 
(1) No matter which dataset we choose, different baselines combined with our strategy can obtain a better consequence in all metrics. 
(2) Take CLINC150 dataset as an example, the error rate of our model drops 8.44\% on BERT relatively, while a 7.77\% reduction on RoBERTa. 
(3) Results demonstrate the availability of our strategy that pays different attention to simple and complex samples throughout training. 

\subsection{Other Curriculum Learning Comparison}

\textbf{Comparison Settings.} We horizontally compared with other CL methods on TNEWS dataset. 
One, refer to \cite{liu2020norm}, sample's difficulty is defined as the norm of 
the eigenvector. The other, \cite{xu2020curriculum} 
uses multiple sub-models for cross-review. According to \cite{xu2020curriculum}, we divide the training 
set into 3 disjoint meta-datasets, train the sub-models separately as teachers, then 
each sample is deduced by other teachers except the one it belongs to. 
We sum the scores as difficulty level. 

\begin{table}[h]
  \caption{Different curriculum learning strategy results.}
  \label{tab:CL_strategy}
  \begin{tabular}{lc}
  \toprule[0.9pt]
  \textbf{Curriculum Learning Strategy} & \textbf{Accuracy}       \\ \midrule[0.9pt]
  BERT                & 55.63    \\ \midrule
  Norm-based CL (ACL2020\shortcite{liu2020norm})                   & 56.02          \\
  Multiple sub-models CL (ACL2020\shortcite{xu2020curriculum})          & 56.07          \\ 
  Density-based CL (\textbf{Ours})   & \textbf{56.55} \\ \bottomrule[0.9pt]
  \end{tabular}
  \end{table}

\textbf{Comparison Results.} The experimental results, displayed in Table \ref{tab:CL_strategy}, 
show that other curriculum learning methods attain a certain 
enhancement of accuracy compared with BERT, as 0.4+\%, while they are slightly weaker than our model, as 0.92\%, 
due to the reason that they fail to exploit the overall distribution information of all samples to define sample's difficulty level simultaneously.  
Besides, their models' capability of defining difficulty level is fixed from the start of training, 
without dynamic renovation. 

\subsection{Validity of Density-Based Difficulty Level}

\subsubsection{Analysis of Error Rate}
\textbf{Comparison Settings.}
Error rate is the proportion of samples that are misclassified in the total samples under 
the same difficulty level. We analyze the sample's error rate under various $K$ difficulty level settings \{2, 3, 4, 7, 10\} on the validation set of TNEWS, which has 2216 misclassified samples with 5000 samples in total. 
Due to the space constraints, we only show the exact number of samples with difficulty level settings in \{2, 10\}. 

\begin{figure}[h]
  \centering
    \includegraphics[width=\linewidth]{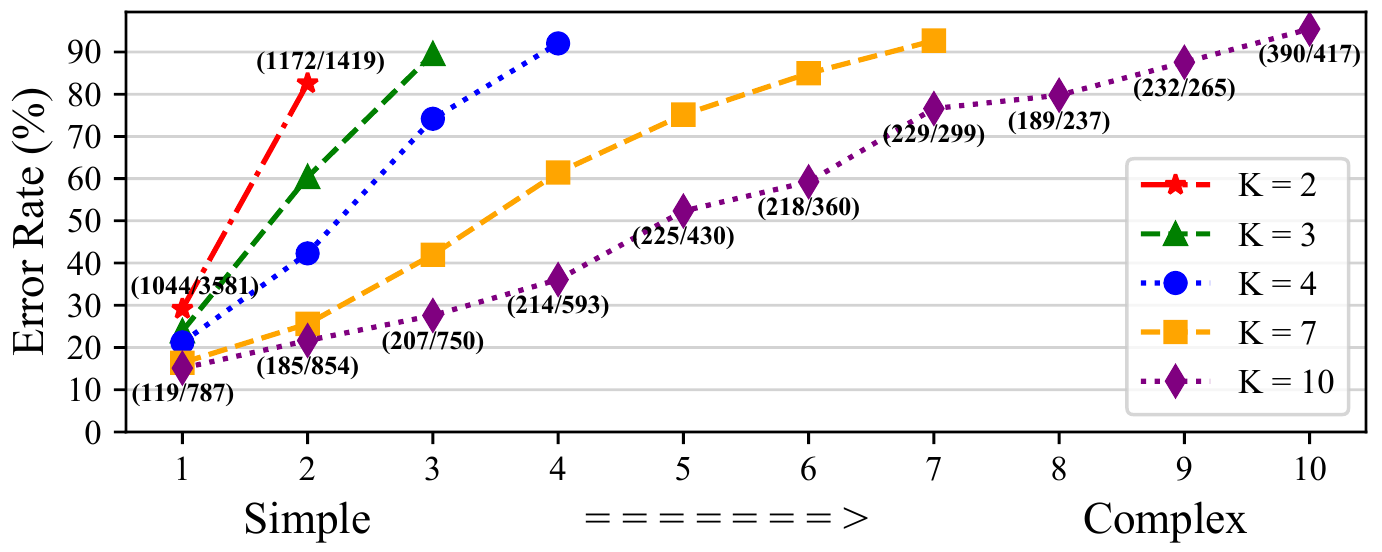}
    \caption{The error rate of samples under various difficulty level settings. For each setting, 
    the error rate is the proportion of samples that are misclassified in the total samples, 
    with the same difficulty level.}
    \label{fig:error-rate}
\end{figure}

\textbf{Comparison Results. }
As Figure \ref{fig:error-rate} depicts: 
(1) More complex samples are incorrectly classified at a much higher rate. 
The error rate of the most complex samples is about 3-6 times above the simplest samples under various level settings. 
(2) For instance, if we divide samples into 10 levels, 
the error rate of simplest samples is 15.12\%, while the error rate of most complex samples is as high as 95.44\%. 
(3) This dulcet observation confirms that our density-based method can distinguish simple and complex samples remarkably.

\subsubsection{Impact of Various Difficulty Level}

\textbf{Comparison Settings.} The total difficulty level number of samples is the number of clusters ($K$) that training samples are separated into. 
The specific cluster operation process is described in Sec. \ref{sec:density}. 
We experiment on TNEWS dataset and set $K$ in
\{2, 3, 4, 7, 10\}.   

\begin{table}[h]
  \caption{Performance on various difficulty level.}
  \label{tab:level}
  \begin{tabular}{p{3.0cm}p{0.60cm}<{\centering}p{0.60cm}<{\centering}p{0.60cm}<{\centering}p{1.26cm}<{\centering}}
  \toprule[0.9pt]
  \textbf{Method}               & \textbf{P}     & \textbf{R}     & \textbf{F1} & \textbf{Accuracy}    \\ \midrule[0.9pt]
  BERT                    & 55.85 & 55.62 & 55.58 & 55.63 \\ \midrule
  BERT+CL (K = 2)     & \textbf{56.92} & 56.37 & 56.39 & 56.37 \\
  BERT+CL (K = 3)     & 56.84 & \textbf{56.55} & \textbf{56.50} & \textbf{56.55} \\
  BERT+CL (K = 4)     & 56.87 & 56.36 & 56.32 & 56.36 \\
  BERT+CL (K = 7)     & 56.65 & 56.14 & 56.12 & 56.14 \\
  BERT+CL (K = 10)    & 56.49 & 56.05 & 56.01 & 56.05 \\ \bottomrule[0.9pt]
  \end{tabular}
  \end{table}

\textbf{Comparison Results.} Table \ref{tab:level} illustrates the corresponding accuracy and other metrics. 
The experimental results express that our model gains higher scores over the BERT model.  
When the difficulty level is set to 3, the model acquires the highest accuracy. 
Though the performance reduces as the difficulty level grows, our model still overmatches the baseline, 
which indicates that our model has certain robustness in the difficulty level settings.

\section{Discussion}

Apart from the pre-trained models, we also experiment on models of the conventional structure.  
Accuracy is the main metric. 
As shown in Table \ref{tab:discuss}, our strategy still outperforms the baselines. 
\begin{table}[h]
  \caption{Performance on models of conventional structure.}
  \label{tab:discuss}
  \begin{tabular}{p{2.0cm}p{1.50cm}<{\centering}p{1.70cm}<{\centering}p{1.20cm}<{\centering}}
  \toprule[0.9pt]
  \textbf{Model}               & \textbf{TNEWS}     & \textbf{BANKING77}     & \textbf{CLINC150}    \\ \midrule[0.9pt]
  Word Average                    & 51.69 & 48.36 & 46.32  \\ 
  +CL (\textbf{Ours})     & \textbf{53.06} & \textbf{50.37} & \textbf{48.09} \\ \midrule
  CNN     & 52.53 & 81.72 & 92.57  \\
  +CL (\textbf{Ours})  & \textbf{53.47} & \textbf{82.44} & \textbf{92.97} \\  \midrule
  LSTM-Attn     & 51.09 & 83.24 & 89.25 \\
  +CL (\textbf{Ours})  & \textbf{51.97} & \textbf{83.84} & \textbf{89.58} \\ \bottomrule[0.9pt]
  \end{tabular}
  \end{table}



\section{Conclusion}

In this work, we propose a \textit{density-based dynamic curriculum learning} model for intent detection. 
Through the density of extracted eigenvectors, the overall 
distribution information of all samples is utilized to define the difficulty 
level simultaneously. With the help of dynamic curriculum learning, model's capability to define difficulty level 
updates adaptively during training. Results indicate 
that our model separate simple and complex samples conspicuously, 
meanwhile, achieve better performance over strong baselines. 

\begin{acks}
The work is supported by National Natural Science
Foundation of China (Grant No.62036001)
and PKU-Baidu Fund (No. 2020BD021).
\end{acks}






\bibliographystyle{ACM-Reference-Format}
\bibliography{sample-base}










\end{document}